\documentclass[twoside,11pt]{article}

\newif\ifarxiv
\arxivtrue

\ifarxiv
\usepackage{jmlr2e_arxiv}
\else
\usepackage{jmlr2e}
\fi
\usepackage{hyperref}

\usepackage{listings}
\newcommand{\innvestigate}{\emph{iNNvestigate}}

\jmlrheading{X}{2018}{XX-XX}{--/--}{--/--}{alber}{Maximilian Alber et. al.}

\ifarxiv
\ShortHeadings{}{}
\else
\ShortHeadings{iNNvestigate}{Maximilian Alber et. al.}
\fi
\firstpageno{1}

\begin{document}

\title{iNNvestigate neural networks!}

\author{%
  \name Maximilian Alber \email maximilian.alber@tu-berlin.de \\
  \addr Technische Universit\"at Berlin, Machine Learning Group\\
       10623 Berlin, Germany
  \AND
  \name Sebastian Lapuschkin \email sebastian.lapuschkin@hhi.fraunhofer.de \\
  \addr Fraunhofer Heinrich Hertz Institute, Video Coding and Analytics \\
       10587 Berlin, Germany
  \AND
  \name Philipp Seegerer \email philipp.seegerer@tu-berlin.de \\
  \name Miriam H\"agele \email haegele@tu-berlin.de \\
  \name Kristof T. Sch\"utt \email kristof.schuett@tu-berlin.de \\
  \name Gr\'egoire Montavon \email gregoire.montavon@tu-berlin.de \\
       \addr Technische Universit\"at Berlin, Machine Learning Group\\
       10623 Berlin, Germany
       \AND
  \name Wojciech Samek \email wojciech.samek@hhi.fraunhofer.de \\
  \addr Fraunhofer Heinrich Hertz Institute, Video Coding and Analytics \\
        10587 Berlin, Germany
  \AND
  \name Klaus-Robert M\"uller \email klaus-robert.mueller@tu-berlin.de \\
       \addr Technische Universit\"at Berlin, Machine Learning Group\\
       10623 Berlin, Germany\\
       \addr Korea University, Department of Brain and Cognitive Engineering\\
       Seoul 02841, Korea\\
       \addr Max Planck Institute for Informatics\\
       66123 Saarbr\"ucken, Germany
       \AND
  \name Sven D\"ahne \email sven.daehne@tu-berlin.de \\
  \name Pieter-Jan Kindermans \email p.kindermans@tu-berlin.de \\
       \addr Technische Universit\"at Berlin, Machine Learning Group\\
       10623 Berlin, Germany
}
\ifarxiv
\editor{}
\else
\editor{TBD}
\fi

\maketitle

\begin{abstract}
  In recent years, deep neural networks have revolutionized many application domains of machine learning and
  are key components of many critical decision or predictive processes.
  Therefore, it is crucial that domain specialists can understand and analyze actions and
  predictions, even of the most complex neural network architectures.
  %
  Despite these arguments neural networks are often treated as black boxes.
  %
  In the attempt to alleviate this shortcoming many analysis methods were proposed,
  yet the lack of reference implementations often makes a systematic comparison between the methods a major effort.
  %
  The presented library \textbf{\innvestigate} addresses this by providing a common interface and out-of-the-box implementation for many analysis methods,
  including the reference implementation for PatternNet and PatternAttribution as well as for LRP-methods.
  To demonstrate the versatility of \innvestigate, we provide an analysis of image classifications for variety of state-of-the-art neural network architectures.
\end{abstract}

\ifarxiv
\else
\fi


\section{Introduction}

In recent years deep neural networks have revolutionized many domains,
e.g., image recognition, speech recognition, speech synthesis, and knowledge discovery~\citep{krizhevsky2012imagenet, lecun2012efficient, schmidhuber2015deep, lecun2015deep, van2016wavenet}.
Due to their ability to naturally learn from structured data and exhibit superior performance,
they are increasingly used in practical applications and critical decision processes,
such as novel knowledge discovery techniques, autonomous driving or medical image analysis. 
To fully leverage their potential it is essential that users can \emph{comprehend and analyze} these processes.
E.g., in neural architecture~\citep{zoph2017learning} or chemical compound searches~\citep{montavon2013machine, schutt2017quantum} it would be extremely useful to know which properties help a neural network to choose appropriate candidates.
Furthermore for some applications understanding the decision process might be a legal requirement.

Despite these arguments neural networks are often treated as black boxes,
because their complex internal workings and the basis for their predictions are not fully understood.
In the attempt to alleviate this shortcoming several methods were proposed, e.g., Saliency Map~\citep{baehrens2010explain, simonyan2013deep}, SmoothGrad~\citep{smilkov2017smoothgrad}, IntegratedGradients~\citep{sundararajan2017axiomatic}, Deconvnet~\citep{zeiler2014visualizing}, GuidedBackprop~\citep{springenberg2015striving}, PatternNet and PatternAttribution~\citep{kindermans2018learning}, LRP~\citep{BachPLOS15, LapCVPR16, LapJMLR16, montavon2018methods}, and DeepTaylor~\citep{MonPR17}.
Theoretically it is not clear which method solves the stated problems best, therefore an empirical comparison is required~\citep{SamTNNLS17, kindermansreliability}.
In order to evaluate these methods, we present \textbf{\innvestigate} which provides a common interface to a variety of analysis methods.

In particular, \innvestigate\ contributes:

\begin{itemize}
\item A common interface for a growing number of analysis  methods that is applicable to a broad class of neural networks.
  With this instantiating a method is as uncomplicated as passing a trained neural network to it and allows for easy qualitative comparisons of methods.
  For quantitative evaluations of (image) classification task we further provide an implementation of the method ``perturbation analysis''~\citep{SamTNNLS17}.
\item Support of all methods listed above---this includes the first reference implementation for PatternNet and PatternAttribution and an extended implementation for LRP---and an open source repository for further contributions.
\item A clean and modular implementation, casting each analysis in terms of layer-wise forward and backward computations. This limits code redundancy, takes advantage of automatic differentiation, and eases future integration of new methods.
\end{itemize}

\innvestigate\ is available at repository: \url{https://github.com/albermax/innvestigate}.
It can be simply installed as Python package and contains documentation for code and applications.
To demonstrate the versatility of \innvestigate\ we provide examples for the analysis of image classifications for a variety of state-of-the-art neural networks.

\paragraph{Terminology}

The different methods pose different assumption to tasks and are designed for different objectives, yet they are related to ``explaining'' or ``interpreting'' neural networks (see~\citet{montavon2018methods}).
We actively refrain from using this terminology in order to prevent misunderstandings between the design choices of the algorithms and the implicit assumption these terms bring along.
Therefore we will solely use the neutral term \textit{analyzing} and leave any interpretation to the user.

\section{Library}



\paragraph{Interface} The main feature is a common interface to several analysis methods.
The workflow is as simple as passing a Keras neural network model to instantiate an analyzer object for a desired algorithm.
Then, if needed, the analyzer will be fitted to the data and eventually be used to analyze the model's predictions.
The corresponding Python code is:

\begin{lstlisting}
import innvestigate
model = create_a_keras_model()
analyzer = innvestigate.create_analyzer("analyzer_name", model)
analyzer.fit(X_train) # if needed
analysis = analyzer.analyze(X_test)
\end{lstlisting}

\paragraph{Implemented methods} At publication time the following algorithms are supported: Gradient Saliency Map, SmoothGrad, IntegratedGradients, Deconvnet, GuidedBackprop, PatternNet and PatternAttribution, DeepTaylor, and LRP including LRP-Z, -Epsilon, -AlphaBeta.
In contrast, current related work \citep{raghakot2017kerasvis, ancona2018towards} is limited to gradient-based methods.
We intend to further extend this selection and invite the community to contribute implementations as new methods emerge.

\begin{figure}[h]
  \centering
  \includegraphics[height=140px]{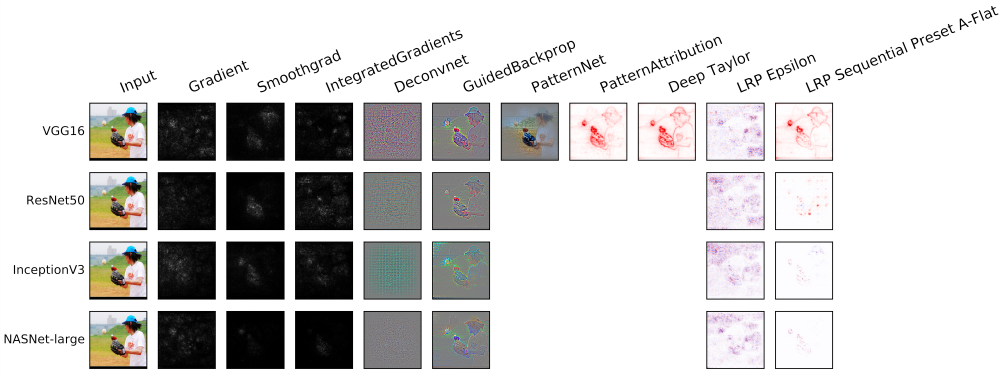}
  \caption{Result of methods applied to various neural networks (blank, if a method does not support a network's architecture yet).}
  \label{figdiffmethods}
\end{figure}

\paragraph{Documentation} The library's documentation contains several introductory scripts and example applications. We demonstrate how the analyses can be applied to the following state-of-the-art models: VGG16 and VGG19~\citep{simonyan2014very}, InceptionV3~\citep{szegedy2016rethinking}, ResNet50~\citep{he2016deep}, InceptionResNetV2~\citep{szegedy2017inception}, DenseNet~\citep{huang2017densely}, NASNet mobile, and NASNet large~\citep{zoph2017learning}. Figure~\ref{figdiffmethods} shows the result of each analysis on a subset of these networks.

\subsection{Details}

\paragraph{Modular implementation} All of the methods have in common that they perform a back-propagation from the model outputs to the inputs.
The core of \innvestigate\ is a set of base classes and functions that is designed to allow for rapid and easy development of such algorithms.
The developer only needs to implement specific changes to the base algorithm and the library will take care of the complex and error-prone handling of the propagation along the graph structure.
Further details can be found in the repositories documentation.

Another advantage of the modular design is that one can extend any analyzer with a given set of wrappers.
One application of this is the smoothing of the analysis results by adding Gaussian noise to the copies of the input and averaging the outcome.
E.g., SmoothGrad is realized in this way by combining a smoothing wrapper with a gradient analyzer.

\paragraph{Training} PatternNet and PatternAttribution~\citep{kindermans2018learning} are two novel approaches that condition their analysis on the data distribution.
This is done by identifying the signal and noise direction for each neuron of a neural network.
Our software scales favorably, e.g., one can train required patterns for the methods on large datasets like Imagenet~\citep{deng2009imagenet} in less than an hour using one GPU.
We present the first reference implementation of these methods.

\paragraph{Quantitative evaluation} Often analysis methods for neural networks are compared by qualitative (visual) inspection of the result.
This is can lead to subjective evaluations and one approach to create a more objective and quantitative comparison of analysis algorithms is the method ``perturbation analysis''~\citep[also known as ``PixelFlipping'']{SamTNNLS17}.
The intuition behind this method is that perturbing regions which are recognized as important for the classification task by the analyzing method, will impact the classification most.
This allows to assess which analysis method best identifies regions that matter for a specific task and neural network.
\innvestigate\ contains an implementation of this method.

\paragraph{Installation \& license} \innvestigate\ is published as open-source software under the MIT-license and can be downloaded from: \url{https://github.com/albermax/innvestigate}.
It is build as a Python 2 or 3 application on top of the popular and established Keras~\citep{chollet2015keras} framework.
This allows to use the library on various platforms and devices like CPUs and GPUs.
At the time of publication only the TensorFlow~\citep{abadi2016tensorflow} Keras-backend is supported.
The library can be simply installed as Python package.

\section{Conclusion}


We have presented \innvestigate, a library that makes it easier to analyze neural networks' predictions and to compare different analysis methods.
This is done by providing a common interface and implementations for many analysis methods as well as making tools for training and comparing methods available.
In particular it contains reference implementations for many methods (PatternNet, PatternAttribution, LRP) and
example application for a large number of state-of-the-art applications.
We expect that this library will support the field of analyzing machine learning and
facilitate research using neural networks in domains such as drug design or medical image analysis.


\acks{Correspondence to MA, SL, KRM, WS and PJK.
This work was supported by the Federal Ministry of Education and Research (BMBF) for the Berlin Big Data Center BBDC (01IS14013A).
Additional support was provided by the BK21 program funded by Korean National Research Foundation grant (No.\ 2012-005741) and the Institute for Information \& Communications Technology Promotion (IITP) grant funded by the Korea government (no.\ 2017-0-00451, No.\ 2017-0-01779).
}


\newpage

\vskip 0.2in
\bibliographystyle{plainnat}
\bibliography{biblio}

\end{document}